\documentclass[runningheads]{llncs}

\usepackage[T1]{fontenc}
\def\doi#1{\href{https://doi.org/\detokenize{#1}}{\url{https://doi.org/\detokenize{#1}}}}
\usepackage{graphicx}
\usepackage{amsmath,amssymb,amsfonts}

\usepackage{algorithm}
\usepackage{algorithmicx}
\usepackage{algpseudocode}
\usepackage{booktabs}
\usepackage[colorlinks=true,linkcolor=blue,citecolor=blue,urlcolor=blue]{hyperref}
\usepackage{subfigure}
\usepackage{arydshln}
\newcommand{\figref}[1]{Fig. \ref{#1}}
\newcommand{\tabref}[1]{Table \ref{#1}}

\usepackage[misc]{ifsym}

\newcommand{\ie}{\textit{i.e.~}}
\newcommand{\eg}{\textit{e.g.~}}

\newcommand{\tsnewidth}{0.225}

\begin{document}
\title{
RandStainNA: Learning Stain-Agnostic Features from Histology Slides by Bridging Stain Augmentation and Normalization
}
\titlerunning{RandStainNA}

\author{
    Yiqing Shen\inst{1}\thanks{Equal contributions. \textsuperscript{\Letter} Correspondence to Jing Ke (kejing@sjtu.edu.cn).}, 
    Yulin Luo\inst{2 \star}, 
    Dinggang Shen\inst{3,4}, 
    Jing Ke\inst{2,5,6}\textsuperscript{(\Letter)}
}
\authorrunning{Y. Shen, Y. Luo, D. Shen, and J. Ke}
\institute{
\textsuperscript{1}School of Mathematical Sciences, Shanghai Jiao Tong University, Shanghai, China\\
\email{shenyq@sjtu.edu.cn}\\
\textsuperscript{2}School of Electronic Information and Electrical Engineering, Shanghai Jiao Tong University, Shanghai, China\\
\email{lyl010221@sjtu.edu.cn}\\
\textsuperscript{3}School of Biomedical Engineering, ShanghaiTech University, Shanghai, China\\
\textsuperscript{4}Shanghai United Imaging Intelligence Co., Ltd., Shanghai, China\\
\email{dgshen@shanghaitech.edu.cn}\\
\textsuperscript{5}School of Computer Science and Engineering, University of New South Wales, Sydney, Australia\\
\textsuperscript{6}BirenTech Research, Shanghai, China\\
\email{kejing@sjtu.edu.cn}\\
}

\maketitle              

\begin{abstract}
Stain variations often decrease the generalization ability of deep learning based approaches in digital histopathology analysis. Two separate proposals, namely stain normalization (SN) and stain augmentation (SA), have been spotlighted to reduce the generalization error, where the former alleviates the stain shift across different medical centers using template image and the latter enriches the accessible stain styles by the simulation of more stain variations. However, their applications are bounded by the selection of template images and the construction of unrealistic styles. To address the problems, we unify SN and SA with a novel RandStainNA scheme, which constrains variable stain styles in a practicable range to train a stain agnostic deep learning model. The RandStainNA is applicable to stain normalization in a collection of color spaces \ie HED, HSV, LAB. Additionally, we propose a random color space selection scheme to gain extra performance improvement. We evaluate our method by two diagnostic tasks \ie tissue subtype classification and nuclei segmentation, with various network backbones. The performance superiority over both SA and SN yields that the proposed RandStainNA can consistently improve the generalization ability, that our models can cope with more incoming clinical datasets with unpredicted stain styles. The codes is available at \url{https://github.com/yiqings/RandStainNA}.

\keywords{Histology Image \and Stain Normalization \and Stain Augmentation.}
\end{abstract}

\section{Introduction}
Pathology visually exams across a diverse range of tissue types obtained by biopsy or surgical procedure under microscopes \cite{histology}. 
Stains are often applied to reveal underlying patterns to increase the contrast between nuclear components and their surrounding tissues \cite{back1}. 
Nevertheless, the substantial variance in each staining manipulation, \eg staining protocols, staining scanners, manufacturers, batches of staining may eventually result in a variety of hue \cite{back2}. 
In contrast to pathologists who have adapted themselves to these variations with years' training, deep learning (DL) methods are prone to suffer from performance degradation, with the existence of inter-center stain heterogeneity \cite{importance}.
Specifically, as color is a salient feature to extract for by deep neural networks, consequently, current successful applications for whole slide images (WSIs) diagnoses are subject to their robustness to color shift among different data centers \cite{importance2}.  
There are two primary directions to reduce the generalization error, namely stain normalization and stain augmentation \cite{sna}.

\begin{figure}[t!]
    \centering
\includegraphics[width=0.8\linewidth]{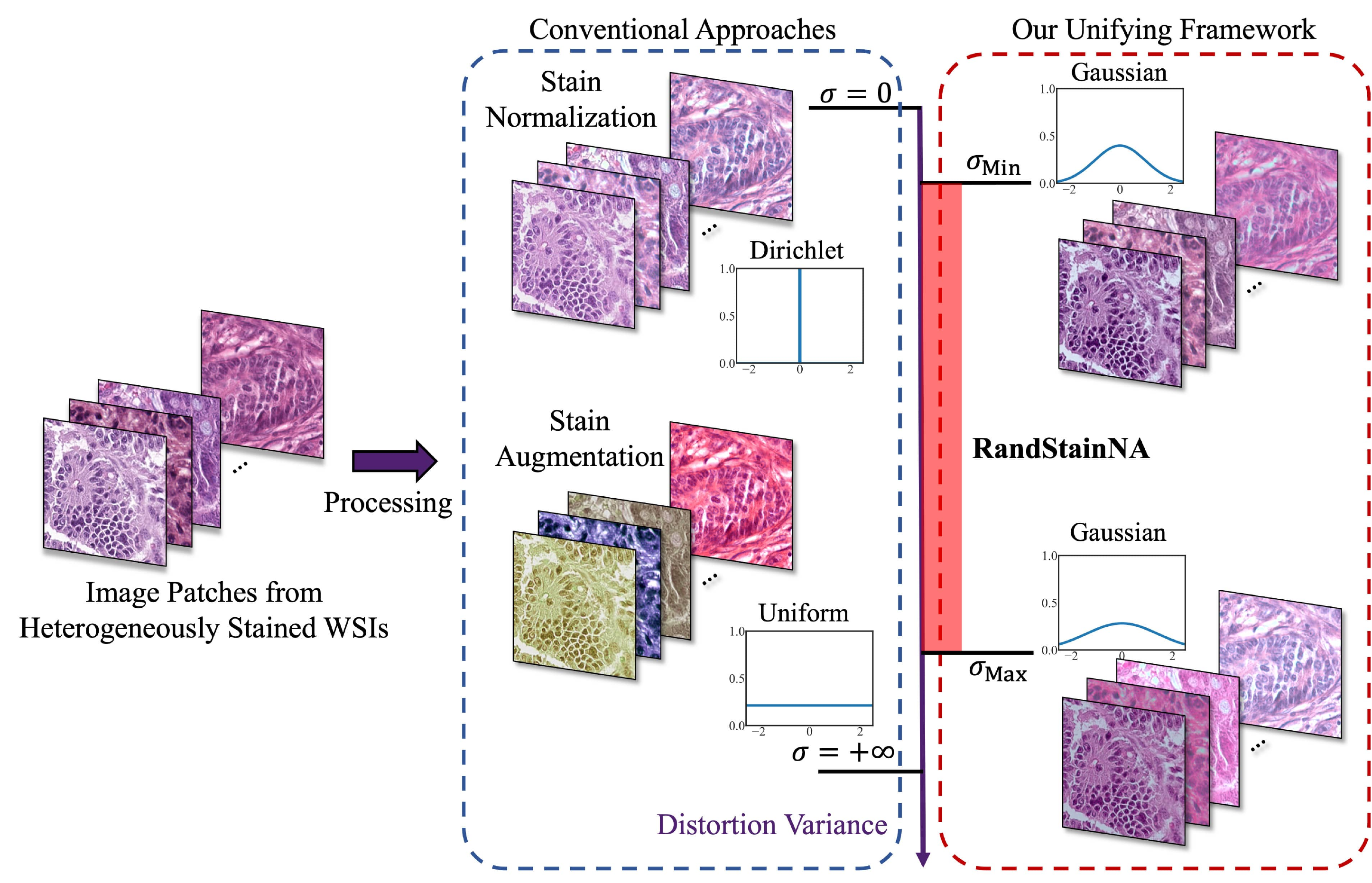}
    \caption{The overall framework of the proposed RandStainNA.}
    \label{fig:intro}
\end{figure}

Stain normalization (SN) aims to reduce the variation by aligning the stain-color distribution of source images to a target template image \cite{sn3,sn2,sn1}. 
Empirical studies regard stain normalization as an essential prerequisite of downstream applications \cite{importance,dataset2,sna}.
Yet, the capability to pinpoint a representative template image for SN relies heavily on domain prior knowledges. 
Moreover, in real-world settings such as federated learning, the template-image selection is not feasible due to the privacy regularizations \cite{back2}, as source images are inaccessible to the central processor as a rule. Some generative adversarial networks (GANs) are proposed recently \cite{sn2,staingan} for SN, yet remaining the phenotype recognizability is always problematic.
A salient drawback of the sole stain style in SN is the restricted color-correlated features can be mined by deep neural networks. Stain augmentation (SA) seeks a converse direction to SN by simulating stain variations while preserving morphological features intact \cite{sa1}. Tellz \textit{et. al.} \cite{uniform} first tailored data augmentations from RGB color space to H\&E color space. Afterward, in parallel with SN approaches, GAN is also widely adopted by stain augmentation applications \eg HistAuGAN \cite{histaugan}.

Previous works have compared the performance between SN and SA without interpretation of their differences \cite{sna}. Moreover, we have observed that the mathematical formulations of SN are coincidental with SA, where the transfer of SN depends on a prior Dirichlet distribution \cite{dirichlet} and SA distorts images with a uniform distribution \cite{uniform}, depicted in \figref{fig:intro}. Hence, we make the first attempt to unify SN and SA for histology image analysis. Two primary contributions are summarized. First, a novel Random Stain Normalization and Augmentation (RandStainNA) method is proposed to bridge stain normalization and stain augmentation, consequently, images can be augmented with more realistic stain styles. Second, a random color space selection scheme is introduced to extend the target scope to various color spaces including HED, HSV, and LAB, to increase flexibility and produce an extra augmentation. The evaluation tasks include tissue classification and nuclei segmentation, and both show our method can consistently improve the performance with a variety of network architectures.

\begin{figure}[t!]
    \centering
    \includegraphics[width=1\linewidth]{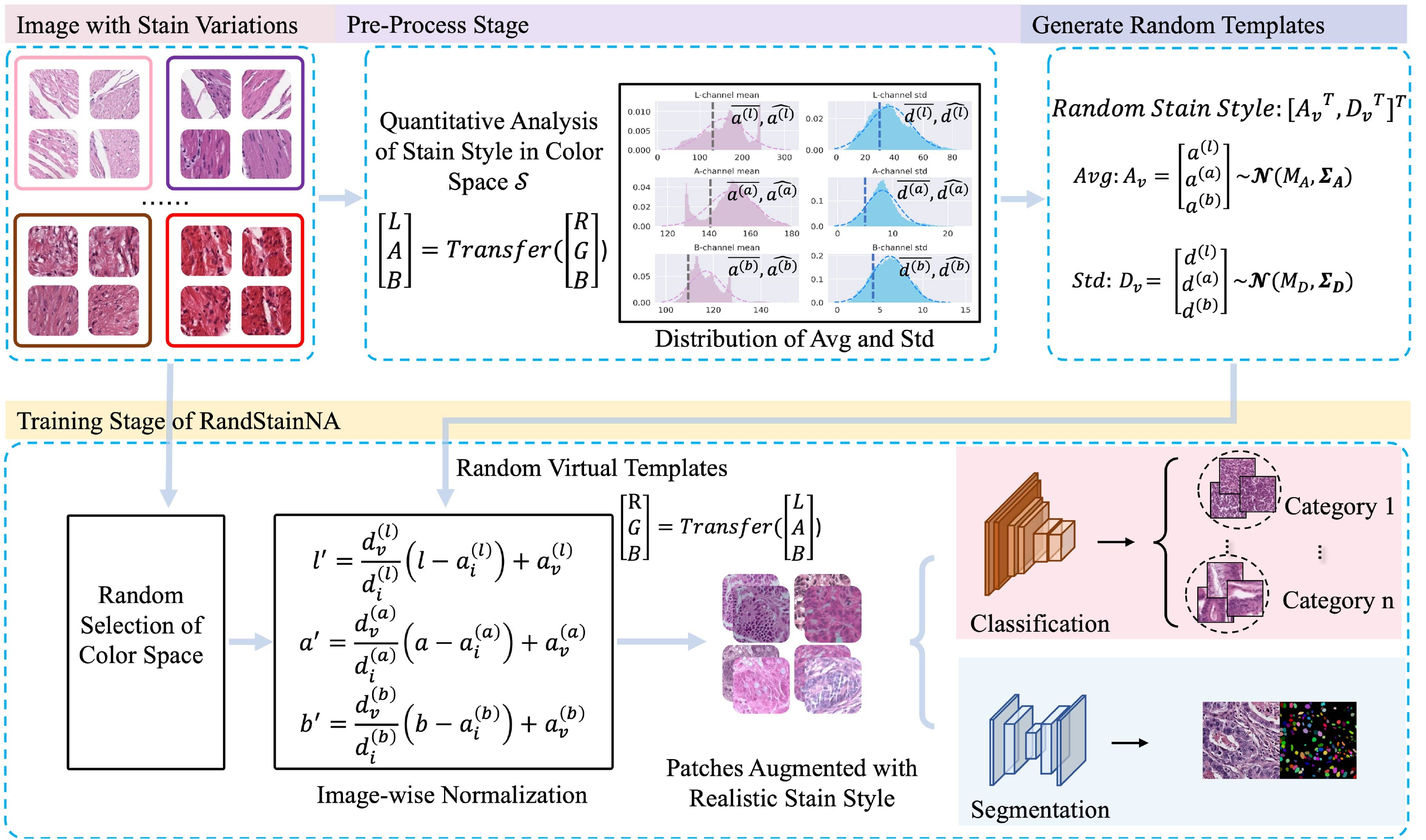}
    \caption{The overall pipeline of the proposed RandStainNA that fuses stain normalization and stain augmentation. Prior to the training stage, random virtual template generation functions are defined \ie $F_\mathbf{M}^\mathcal{S}=\mathcal{N}(\mathbf{M}_A^\mathcal{S},\mathbf{\Sigma}_A^\mathcal{S})$ and $F_\mathbf{D}^\mathcal{S}=\mathcal{N}(\mathbf{M}_D^\mathcal{S},\mathbf{\Sigma}_D^\mathcal{S})$. The three-step training stage comprises a random selection of color space $\mathcal{S}$, the generated of an associated random stain style template  $[\mathbf{M}_v^\mathcal{S},\mathbf{\Sigma}_v^\mathcal{S}]$, and the normalization of a batch with the generated virtual template. Our approach is downstream task agnostic. } 
    \label{fig:overall}
\end{figure}

\section{Methodology}
\subsubsection{Method Overview.} \textbf{Rand}om \textbf{Stain} \textbf{N}ormalization and \textbf{A}ugmentation (RandStainNA) is a hybrid framework designed to fuse stain normalization and stain augmentation to generate more realistic stain variations. It incorporates randomness to SN by automatically sorting out a random virtual template from pre-estimated stain style distributions. More specifically, from the perception of SN's viewpoint, stain styles `visible' to the deep neural network are enriched in the training stage. Meanwhile, from the perception from the SA's viewpoint, RandStainNA imposes a restriction on the distortion range and consequently, only a constrained practicable range is `visible' to CNN. The framework is a general strategy and task agonist, as depicted in \figref{fig:overall}.

\subsubsection{Stain Style Creation and Characterization.}
Unlike the formulation of comprehensive color styles of nature images, stain style of histology remains to be a vague concept, which is primarily based on visual examination, restricting to obtain a precise objective for alleviating the stain style variation \cite{reinhard,sn1}. To narrow this gap, our work first qualitatively defines the stain style covering six parameter, namely the average and standard deviation of each channel in LAB color space \cite{reinhard}. We pick up LAB space for its notable capability to represent heterogeneous styles in medical images \cite{reinhard}. Novelly, we first transfer all histology slides in the training set from RGB space to LAB color space. Then the stain style of image $\mathbf{x}_i$ are depicted by $\mathbf{A}_i=[a^{(l)}_i,a^{(a)}_i,a^{(b)}_i]\in\mathbb{R}^3$ and $\mathbf{D}_i=[d^{(l)}_i,d^{(a)}_i,d^{(b)}_i]\in\mathbb{R}^3$, where $a^{(c)}_i,d^{(c)}_i$ are the average value and standard deviation of each channel $c\in\{l,a,b\}$ in image $\mathbf{x}_i$, as shown in the pre-processing stage block in \figref{fig:overall}. 

\subsubsection{Virtual Stain Normalization Template.}
In routine stain normalization approaches \cite{reinhard,sn1}, a source image is normalized to a pre-selected template image by aligning the average of $\mathbf{A}_s$ and standard deviation $\mathbf{D}_s$ of pixel values to the template $\mathbf{A}_t$ and $\mathbf{D}_t$. Thus, it is sufficient to formulate a template image with $[\mathbf{A}_t,\mathbf{D}_t]$. In the proposed RandStainNA, we expand the uniformly shared one-template mode to a board randomly generated virtual templates $[\mathbf{A}_v,\mathbf{D}_v]$ scheme. To be more specific, iteratively, random $\mathbf{A}_v$ is sampled from distribution $F_\mathbf{A}$, and likewise $\mathbf{D}_v$ is picked out from the other distribution $F_\mathbf{D}$, which are jointly used as the target for every training sample normalization. Empirical results yield that eventual performance are robust to a wide range of distribution types of $F_\mathbf{A}$ and $F_\mathbf{D}$, such as Gaussian and t-distribution. In the rest of this section, we simply leverage Gaussian distribution as the estimation \ie setting $F_\mathbf{A} = \mathcal{N}(\mathbf{M}_A,\mathbf{\Sigma}_A)$, $F_\mathbf{D} = \mathcal{N}(\mathbf{M}_D,\mathbf{\Sigma}_D)$, where $\mathcal{N}(\mathbf{M},\mathbf{\Sigma})$ writes for the Gaussian distribution with expectation $\mathbf{M}$ and covariance matrix $\mathbf{\Sigma}$. Notably, due to the orthogonality of channels, $\mathbf{\Sigma}$ is a diagonal matrix \ie $\mathbf{\Sigma} = \operatorname{diag}(\sigma_1^2,\sigma_2^2,\sigma_3^2)$ for some $\sigma_j$ with $j=1,2,3$.

\subsubsection{Statistics Parameters Estimation For Virtual Template Generation.} The estimation of statistical parameters of $\mathbf{M}_A,\mathbf{\Sigma}_A,\mathbf{M}_D,\mathbf{\Sigma}_D$ are afterwards applied to the formation of stain style discussed above. A proper candidate is attributed to the sample channel mean values of all the training images for $\mathbf{M}_A $ and $\mathbf{M}_D$, as well as the standard deviations of samples for $\mathbf{\Sigma}_A $ and $ \mathbf{\Sigma}_D$, based on the average value and standard deviation of the whole training set. However, two defects turn out in this discipline that one is the inefficiency to transverse the whole set, and the other is the special cases of infeasibility \eg federated learning or lifelong learning. Therefore, we provide a more computation-efficient alternative, by randomly curating a small number of patches from the training set and applying their sample mean and standard deviation as $\mathbf{M}_A,\mathbf{\Sigma}_A,\mathbf{M}_D,\mathbf{\Sigma}_D$. The empirical results suggest it can achieve competitive performance.

\subsubsection{Image-wise Normalization With Random Virtual Template.} After transferring image $\mathbf{x}_i$ from RGB into LAB space, we write the pixel value as $[l,a,b]$. We denote the average and standard deviation (std) for each channel of this image as $\mathbf{A}_i=[a^{(l)}_i,a^{(a)}_i,a^{(b)}_i]$ and $\mathbf{D}_i=[d^{(l)}_i,d^{(a)}_i,d^{(b)}_i]$, and the generated random virtual template associated to $\mathbf{x}_i$ from $F_\mathbf{A},F_\mathbf{D}$ as $\mathbf{A}_v=[a^{(l)}_v,a^{(a)}_v,a^{(b)}_v]$ and $\mathbf{D}_i=[d^{(l)}_v,d^{(a)}_v,d^{(b)}_v]$. Then the image-wise normalization based on random virtual template is formulated as
\begin{equation}
    \begin{cases}
    \begin{aligned}
    l^\prime &= \frac{d^{(l)}_v}{d^{(l)}_i}(l - a^{(l)}_i) + a^{(l)}_v \\
    a^\prime &= \frac{d^{(a)}_v}{d^{(a)}_i}(a - a^{(a)}_i) + a^{(a)}_v \\
    b^\prime &= \frac{d^{(b)}_v}{d^{(b)}_i}(b - a^{(b)}_i) + a^{(b)}_v,
    \end{aligned}
    \end{cases}
\end{equation}
Then, we transfer $[l^\prime,a^\prime,b^\prime]$ from LAB back to RGB space. Notably, we generate different virtual templates for images that vary at every epoch during the training stage. Therefore, RandStainNA can largely increase the data variations with the on-the-fly generate virtual templates.

\subsubsection{Determine Random Color Space for Augmentation.} By the computation of the stain style parameters $[\mathbf{M}, \mathbf{\Sigma}]$ of distinct color spaces, we can derive their associated $F_\mathbf{A}$ and $F_\mathbf{D}$. Afterwards, we extend our RandStainNA from LAB to other color spaces \eg HED, HSV. This extension allows the proposal of a random color space selection scheme, which will further strengthen the regularization effect. The candidate pool comprises three widely-used color spaces in the domain of histology, \ie HED, HSV, LAB. Training iteratively, an initial color space $\mathcal{S}$ is an arbitrary decision with equal probability \ie $p=\frac{1}{3}$, or with manually-assigned values depending of the performance of each independent space. Subsequently, a virtual template is assigned to associate with $\mathcal{S}$ to perform image-wise stain normalization.

\section{Experiments}

\subsubsection{Dataset and Evaluation Metrics.}
We evaluate our proposed RandStainNA on two image analysis tasks \ie classification and segmentation. Regarding the patch-level classification task, we use a widely-used histology public dataset NCT-CRC-HE-100K-NONORM for training and validation, with the addition of the CRC-VAL-HE-7K dataset for external testing \cite{dataset1}. These two sets comprise a number of 100,000 and 7180 histological patches respectively, from colorectal cancer patients of multiple data centers with heterogeneous stain styles. We randomly pick up 80\% from NCT-CRC-HE-100K-NONORM for training and the rest 20\% for validation. The original dataset covers nine categories, but for the category of background, we can straightforwardly identify them in pre-processing stage with OTSU algorithm \cite{otsu} and thus it is removed in our experiment for a more reliable result. The top-1 classification accuracy (\%) is used as the metric for the 8-category classification task. For the nuclei segmentation task, we use a small public dataset of MoNuSeg\cite{dataset2}, with Dice and IoU as the metrics.

\begin{table}[!t]
\centering
\caption{Test accuracy (\%) comparison on the tissue type classification task. We compare our method with stain augmentation (SA) \cite{sna} and stain normalization (SN) \cite{reinhard} in three color space \ie LAB, HSV, HED \cite{sna}. In SA, we follow previous work \cite{sna} by leveraging two settings, namely light (L) and strong (S), determined by the degree of distortion. The best and second best are marked in boldface and with * respectively.} 
\label{tab:com}
\resizebox{\linewidth}{!}{ 
\begin{tabular}{l|c|c|c|c|c|c} 
\toprule
Method & 
ResNet18 & ResNet50 & MobileNet & EfficientNet & ViT & SwinTransformer \\
\hline
Baseline & 84.20 & 72.07 & 80.25 & 79.62 & 72.85 & 71.06 \\
\hline
SA1-L+LAB & 84.62 & 77.24 & 84.09 & 81.71 & 75.17 & 69.42 \\
SA1-S+LAB & 89.35 & 87.97 & 90.79 & 90.81 & 84.77 & 76.76 \\
SA2-L+LAB & 93.55 & 92.81 & 92.72 & 93.66 & 89.90 & 88.42 \\
SA2-S+LAB & 90.77 & 90.50 & 89.13 & 89.64 & 84.84 & 75.13 \\
SA1-L+HED & 92.47 & 89.55 & 90.81 & 92.57 & 86.22 & 81.17 \\
SA1-S+HED & 88.77 & 87.92 & 86.58 & 88.67 & 87.42 & 76.50 \\
SA2-L+HSV & 91.39 & 88.97 & 88.97 & 91.10 & 82.80 & 78.70 \\
SA2-S+HSV & 91.93 & 90.06 & 90.22 & 91.76 & 85.81 & 77.83 \\
SN+LAB & 93.01 & 91.40 & 92.23 & 91.92 & 89.77 & 88.80 \\
SN+HED & 91.38 & 90.57 & 89.54 & 91.34 & 88.29 & 86.60 \\
SN+HSV & 93.85 & 93.86 & 92.38 & 93.90 & 90.63 & 86.21 \\
\hline
Ours (LAB) & 94.44* & 93.97 & 93.94 & 93.54 & 90.30 & 91.01 \\
Ours (HED) & 93.28 & 93.61 & 91.69 & 92.67 & 91.41 & 90.03 \\
Ours (HSV) & 94.04 & 94.12* & 94.06* & \textbf{94.81} & 93.27* & \textbf{92.75} \\
\hdashline
Ours (Full) & \textbf{94.66} & \textbf{94.45} & \textbf{94.53} & 94.62* & \textbf{93.34} & 92.39* \\
\bottomrule
\end{tabular}
}
\vspace{1em}
\caption{Performance comparison on nuclei segmentation in terms of Dice and IoU.}\label{tab:seg}
\resizebox{\linewidth}{!}{ 
    \begin{tabular}{l|cccccccc} 
    \toprule
    Metrics& Baseline & SA+LAB & SA+HED & SA+HSV & SN+LAB & SN+HED & SN+HSV & Ours \\
    \hline
     Dice & 0.7270 & 0.7297 & 0.7354 & 0.7349 & 0.7792 & 0.7668 & 0.7780 & \textbf{0.7802} \\
    IoU & 0.5564 & 0.5665 & 0.5758 & 0.5712 &  0.6302 & 0.6119 & 0.6291 & \textbf{0.6335}  \\
    \bottomrule
    \end{tabular}
}
\end{table}

\begin{figure}[h!]
    \centering
    \includegraphics[width=1\linewidth]{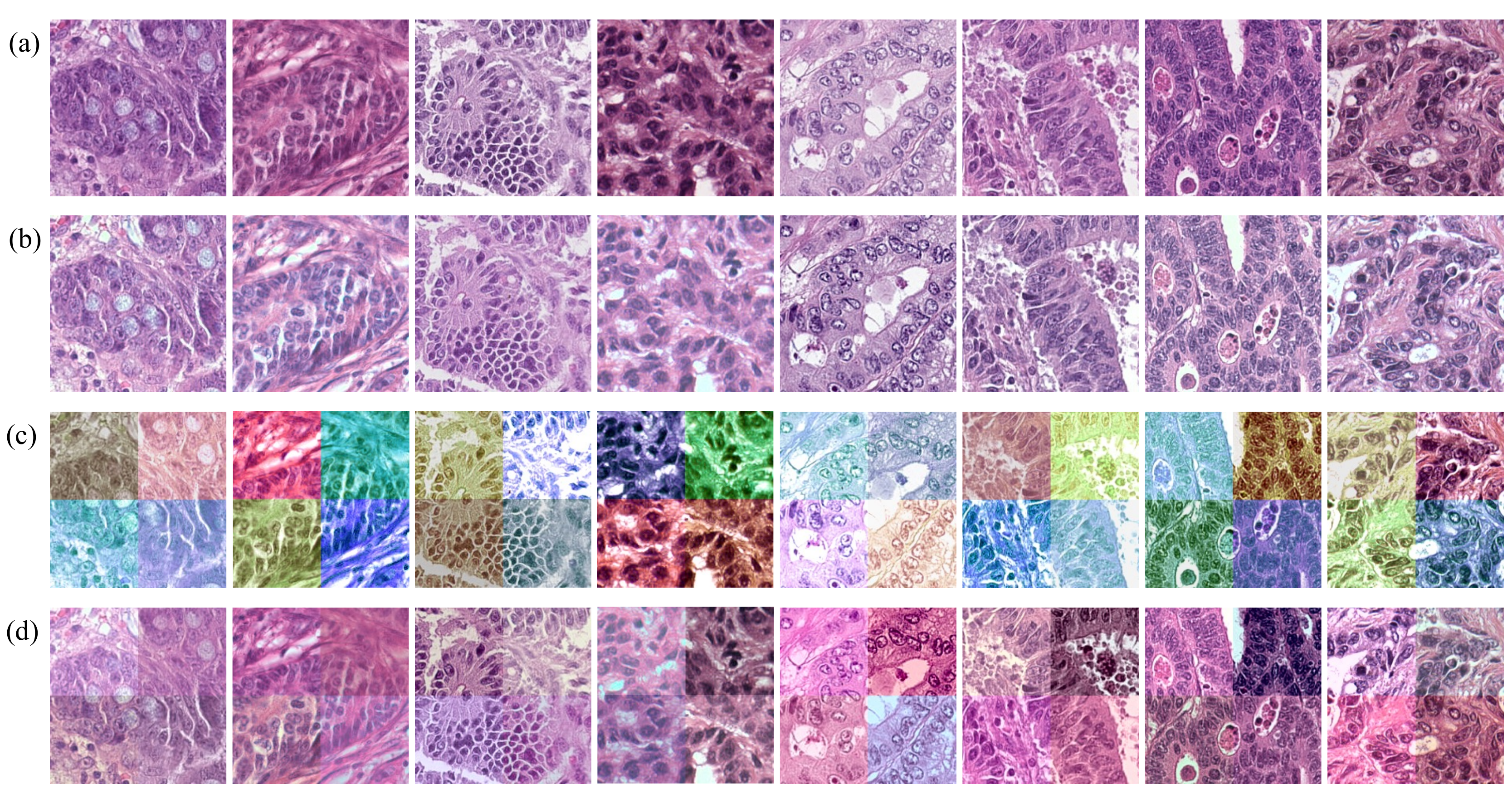}
    \caption{The illustrative patch examples of (a) raw images, (b) stain-normalized images, (c) stain-augmented images, (d) images processed with the proposed RandStainNA. We incorporate the results of four random runs into one image patch to demonstrate the different grades of randomness maybe achieved by the stain augmentation methods and our RandStainNA in (c) and (d).} 
    \label{fig:visualization}
\includegraphics[width=0.96\linewidth]{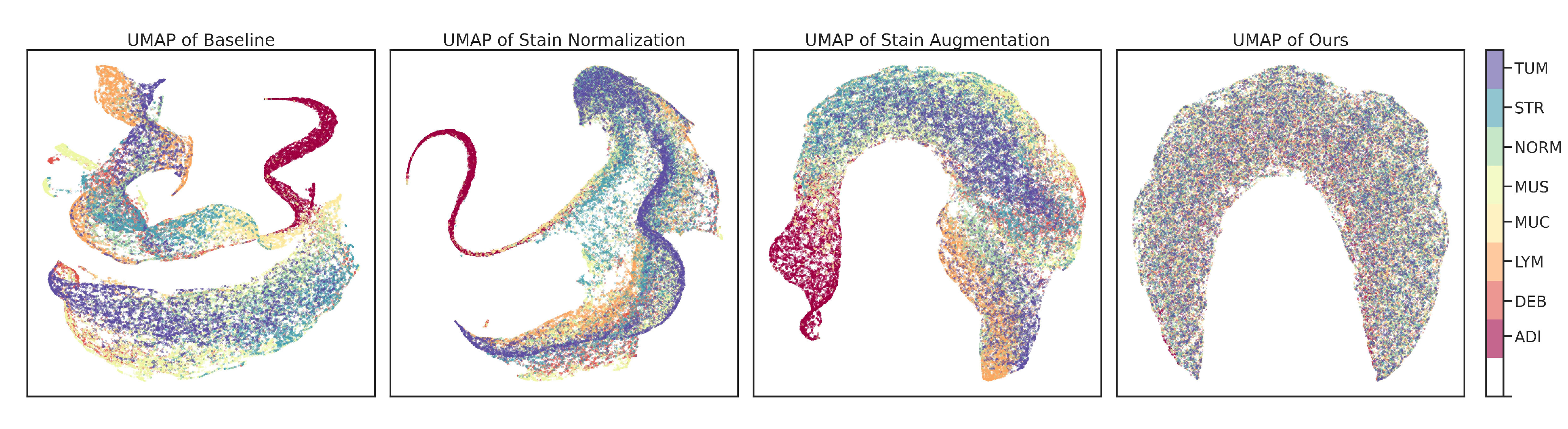}
\caption{UMAP \cite{umap} embedding of the stain style charactiersitic statistics \ie $[\mathbf{M},\mathbf{\Sigma}]$ of raw images, stain normalized images, stain augmented images and those augmented with our RandStainNA. As shown, our method can enrich the realistic stain styles in training CNNs.}
\label{fig:umap}
\end{figure}

\subsubsection{Network Architecture and Settings.}
In the classification task, we employ six backbone architectures to perform the evaluations, namely the ResNet-18 \cite{resnet}, ResNet-50 \cite{resnet}, MobileNetV3-Small \cite{mobilenet}, EfficientNetB0 \cite{efficientnet}, ViT-Tiny \cite{vit} and SwinTransformer-Tiny \cite{swin}. These networks, including CNN and transformers, may represent a wide range of network capabilities, which effectively demonstrate the adaptability of our method in different settings. In the nuclei segmentation task, we use CIA-Net as the backbone \cite{cianet} for its notable performance in small set processing. We use a consistent training scheme for distinct networks for performance comparison with stain augmentation and stain normalization methods. Detailed training schemes and hyper-parameter settings are shown in the supplementary material. We perform 3 random runs and compute the average for each experiments.

\subsubsection{Compared Methods.}
All models are trained with morphology augmentation, namely the random vertical and horizontal flip. In both evaluation tasks, we compare our method with existing stain normalization \cite{reinhard} and stain augmentation \cite{sa1,sna} approaches performed in the three color spaces \ie HED, HSV, LAB. Regarding the stain augmentation in HED, we employ a multiplication rule \cite{sna} that adds noise to each channel \ie $p^\prime = p * \varepsilon_1 + \varepsilon_2$, where $p^\prime$ is the augmented pixel value and $p$ is the original pixel value, and $\varepsilon_1 $ and $\varepsilon_2$ are uniform random noises, termed as stain augmentation scheme \#1 (SA1). For the SA in HSV, we adopt an addition rule \ie $p^\prime = p+ p * \varepsilon$ \cite{sna}, termed as stain augmentation scheme \#2 (SA2). We integrate the above two schemes for LAB stain augmentation, due to an absence of literature works for SA in LAB. We also configure two augmentation settings according to different degrees of distortion \ie range of random noise, denoted as light (L) and strong (S) \cite{sa1,sna}. To fully retain recognizable morphological features, we do not take GAN-related approaches for comparison.

\subsubsection{Results.}
Our method can consistently improve the baseline performance of the six backbone architectures in terms of test accuracy, with the implementation in three color spaces, as shown in \tabref{tab:com}. Therefore, it yields the effectiveness of RandStainNA. The hybrid architecture can outperform a sole deployment of either SN or SA. With the random color space selection scheme (denoted as `full'), the RandStainNA achieves further performance improvement. To demonstrate the effects achieved with different approaches straightforwardly, we visualize the original raw images with stain variations, SN images, SA images, and images processed with our RandStainNA in \figref{fig:visualization}. In the visualization graph, we use the results from SA and SN performed in HSV space as an example, which shows very similar outcomes in LAB and HED spaces. As shown, the SN unifies stain styles into a shared template that may leave out many useful features \cite{sna}, and the SA may generate unrealistic images. In contrast, our method generates much more realistic images to reorganize by both human and deep learning techniques. The \figref{fig:umap} provides the UMAP embedding of stain style parameters of $[\mathbf{M},\mathbf{\Sigma}]$ in the associated solutions. 
The nuclei segmentation results are listed in the \tabref{tab:seg}. For SN and SA in each color space, we pick up one configuration with higher performance in the classification task. Our method also achieves the best performance to demonstrate its effectiveness in various downstream tasks.

\subsubsection{Ablation Study.} 
The ablation study is performed on the classification task. First, we test the effect of the distribution style of $F_\mathbf{A},F_\mathbf{D}$. The test accuracy is 93.98, 93.90, 93.04, 92.48 for Gaussian, t-distribution, uniform, and Laplace respectively. The effect of sample numbers to compute the sample mean and the standard deviation is also evaluated. The test accuracies are 93.42, 93.29, 94.08, 93.98 for the cases of computing the averages and standard deviations with 10 images per category, 100 images per category, 1000 images per category, and the whole training set respectively, which yields the robustness to $\mathbf{M}$ and $\mathbf{\Sigma}$.

\section{Conclusion}

The proposed RandStainNA framework aims to cope with the inevitable stain variance problem for clinical pathology image analysis. Leveraging the advantages of both stain normalization and stain augmentation, the proposed framework produces more realistic stain variations to train stain agnostic DL models. Additionally, RandStainNA is straightforward practically and efficient when applied as an on-the-fly augmentation technique, in comparison with most current GANs. Moreover, the result shows the feasibility to train robust downstream classification and segmentation networks on various architectures. One future direction of our current works is the expansion of color spaces, \eg YUV, YCbCr, YPbPr, YIQ, XYZ \cite{colorspace}, to further improve the generalization ability.

\subsubsection{Acknowledgements.} This work has been supported by NSFC grants 62102247.

\bibliographystyle{splncs04}
\bibliography{ref}

\begin{thebibliography}{10}
\providecommand{\url}[1]{\texttt{#1}}
\providecommand{\urlprefix}{URL }
\providecommand{\doi}[1]{https://doi.org/#1}

\bibitem{umap}
Becht, E., et~al.: Dimensionality reduction for visualizing single-cell data
  using umap. Nature biotechnology  \textbf{37}(1),  38--44 (2019)

\bibitem{importance}
Ciompi, F., et~al.: The importance of stain normalization in colorectal tissue
  classification with convolutional networks. In: 2017 IEEE 14th International
  Symposium on Biomedical Imaging (ISBI 2017). pp. 160--163. IEEE (2017)

\bibitem{vit}
Dosovitskiy, A., et~al.: An image is worth 16x16 words: Transformers for image
  recognition at scale. arXiv preprint arXiv:2010.11929  (2020)

\bibitem{colorspace}
Gowda, S.N., et~al.: Colornet: Investigating the importance of color spaces for
  image classification. In: Asian Conference on Computer Vision. pp. 581--596.
  Springer (2018)

\bibitem{importance2}
Gupta, V., et~al.: Automated classification for breast cancer histopathology
  images: Is stain normalization important? In: Computer Assisted and Robotic
  Endoscopy and Clinical Image-Based Procedures, pp. 160--169. Springer (2017)

\bibitem{histology}
Gurcan, M.N., et~al.: Histopathological image analysis: A review. IEEE reviews
  in biomedical engineering  \textbf{2},  147--171 (2009)

\bibitem{resnet}
He, K., et~al.: Deep residual learning for image recognition. In: Proceedings
  of the IEEE conference on computer vision and pattern recognition. pp.
  770--778 (2016)

\bibitem{mobilenet}
Howard, A., et~al.: Searching for mobilenetv3. In: Proceedings of the IEEE/CVF
  International Conference on Computer Vision. pp. 1314--1324 (2019)

\bibitem{dataset1}
Kather, J.N., et~al.: Predicting survival from colorectal cancer histology
  slides using deep learning: A retrospective multicenter study. PLoS medicine
  \textbf{16}(1),  e1002730 (2019)

\bibitem{back2}
Ke, J., et~al.: Style normalization in histology with federated learning. In:
  2021 IEEE 18th International Symposium on Biomedical Imaging (ISBI). pp.
  953--956. IEEE (2021)

\bibitem{sn3}
Khan, A.M., et~al.: A nonlinear mapping approach to stain normalization in
  digital histopathology images using image-specific color deconvolution. IEEE
  Transactions on Biomedical Engineering  \textbf{61}(6),  1729--1738 (2014)

\bibitem{dataset2}
Kumar, N., et~al.: A multi-organ nucleus segmentation challenge. IEEE
  transactions on medical imaging  \textbf{39}(5),  1380--1391 (2019)

\bibitem{swin}
Liu, Z., et~al.: Swin transformer: Hierarchical vision transformer using
  shifted windows. In: Proceedings of the IEEE/CVF International Conference on
  Computer Vision. pp. 10012--10022 (2021)

\bibitem{back1}
Nadeem, S., et~al.: Multimarginal wasserstein barycenter for stain
  normalization and augmentation. In: International Conference on Medical Image
  Computing and Computer-Assisted Intervention. pp. 362--371. Springer (2020)

\bibitem{otsu}
Otsu, N.: A threshold selection method from gray-level histograms. IEEE
  transactions on systems, man, and cybernetics  \textbf{9}(1),  62--66 (1979)

\bibitem{reinhard}
Reinhard, E., et~al.: Color transfer between images. IEEE Computer graphics and
  applications  \textbf{21}(5),  34--41 (2001)

\bibitem{sn2}
Salehi, P., et~al.: Pix2pix-based stain-to-stain translation: a solution for
  robust stain normalization in histopathology images analysis. In: 2020
  International Conference on Machine Vision and Image Processing (MVIP).
  pp.~1--7. IEEE (2020)

\bibitem{staingan}
Shaban, M.T., et~al.: Staingan: Stain style transfer for digital histological
  images. In: 2019 Ieee 16th international symposium on biomedical imaging
  (Isbi 2019). pp. 953--956. IEEE (2019)

\bibitem{efficientnet}
Tan, M., et~al.: Efficientnet: Rethinking model scaling for convolutional
  neural networks. In: International conference on machine learning. pp.
  6105--6114. PMLR (2019)

\bibitem{uniform}
Tellez, D., et~al.: H and e stain augmentation improves generalization of
  convolutional networks for histopathological mitosis detection. In: Medical
  Imaging 2018: Digital Pathology. vol. 10581, p. 105810Z. International
  Society for Optics and Photonics (2018)

\bibitem{sa1}
Tellez, D., et~al.: Whole-slide mitosis detection in h\&e breast histology
  using phh3 as a reference to train distilled stain-invariant convolutional
  networks. IEEE transactions on medical imaging  \textbf{37}(9),  2126--2136
  (2018)

\bibitem{sna}
Tellez, D., et~al.: Quantifying the effects of data augmentation and stain
  color normalization in convolutional neural networks for computational
  pathology. Medical image analysis  \textbf{58},  101544 (2019)

\bibitem{histaugan}
Wagner, S.J., et~al.: Structure-preserving multi-domain stain color
  augmentation using style-transfer with disentangled representations. In:
  International Conference on Medical Image Computing and Computer-Assisted
  Intervention. pp. 257--266. Springer (2021)

\bibitem{sn1}
Wang, Y.Y., et~al.: A color-based approach for automated segmentation in tumor
  tissue classification. In: 2007 29th Annual International Conference of the
  IEEE Engineering in Medicine and Biology Society. pp. 6576--6579. IEEE (2007)

\bibitem{dirichlet}
Zanjani, F.G., et~al.: Stain normalization of histopathology images using
  generative adversarial networks. In: 2018 IEEE 15th International symposium
  on biomedical imaging (ISBI 2018). pp. 573--577. IEEE (2018)

\bibitem{cianet}
Zhou, Y., et~al.: Cia-net: Robust nuclei instance segmentation with
  contour-aware information aggregation. In: International Conference on
  Information Processing in Medical Imaging. pp. 682--693. Springer (2019)

\end{thebibliography}

\newpage
\section*{Supplementary Material}

\begin{table}[h!]
\centering
\caption{Test F1-score comparison on the tissue type classification task. We follow the same settings as Table 1 in paper. The best performance is marked in boldface. Our proposed consistently achieve the best performance.} 
\label{tab:com}
\resizebox{\linewidth}{!}{ 
\begin{tabular}{l|c|c|c|c|c|c} 
\toprule
Method & 
ResNet18 & ResNet50 & MobileNet & EfficientNet & ViT & SwinTransformer \\
\hline
Baseline & 0.785 & 0.628 & 0.737 & 0.738 & 0.673 & 0.624 \\
\hline
SA1-L+LAB & 0.784 & 0.689 & 0.788 & 0.761 & 0.699 & 0.594 \\
SA1-S+LAB & 0.853 & 0.815 & 0.870 & 0.869 & 0.795 & 0.691 \\
SA2-L+LAB & 0.913 & 0.904 & 0.901 & 0.914 & 0.863 & 0.845 \\
SA2-S+LAB & 0.877 & 0.875 & 0.859 & 0.864 & 0.817 & 0.690 \\
SA1-L+HED & 0.890 & 0.858 & 0.875 & 0.897 & 0.831 & 0.764 \\
SA1-S+HED & 0.853 & 0.839 & 0.831 & 0.855 & 0.845 & 0.699 \\
SA2-L+HSV & 0.883 & 0.854 & 0.855 & 0.880 & 0.786 & 0.718 \\
SA2-S+HSV & 0.896 & 0.866 & 0.876 & 0.892 & 0.825 & 0.705 \\
SN+LAB & 0.899 & 0.870 & 0.883 & 0.878 & 0.855 & 0.835 \\
SN+HED & 0.890 & 0.860 & 0.852 & 0.879 & 0.838 & 0.800 \\
SN+HSV & 0.920 & 0.922 & 0.900 & 0.921 & 0.885 & 0.805 \\
\hline
Ours (LAB) & \textbf{0.927} & 0.912 & 0.922 & 0.914 & 0.857 & 0.873 \\
Ours (HED) & 0.905 & 0.915 & 0.881 & 0.904 & 0.898 & 0.866 \\
Ours (HSV) & 0.920 & 0.922 & 0.920 & \textbf{0.930} & 0.913 & \textbf{0.899} \\
\hdashline
Ours (Full) & 0.926 & \textbf{0.925} & \textbf{0.927} & 0.926 & \textbf{0.915
} & 0.896 \\
\bottomrule

\end{tabular}
}
\end{table}

\begin{table}[h]
\centering
\caption{Training configurations. Notably, for all the compared methods \ie stain normalization and stain augmentation, we use a consistent training settings. E, BS, LR, Opt., M and WD represents the epoch, batch size, learning rate, optimization, momentum and weight decay ratio respectively. }\label{tab:training_settings}
    \begin{tabular}{l|c|c|c|c|c|c|c|c} 
    \toprule
    Networks & E & BS & LR & Opt. & M & WD & Scheduler & Other Strategy \\
    \hline 
    ResNet18 & 50 & 128 & 1e-1$\to$1e-3  & SGD & 0.9 & 1e-4 & Cosine & Warm-up w. 3ep \\
    ResNet50 & 50 & 128 & 1e-1$\to$1e-3 & SGD & 0.9 & 1e-4 & Cosine & Warm-up w. 3ep \\
    MobileNetV3-S & 50 & 128 & 1e-1$\to$1e-3 & SGD & 0.9 & 1e-4 & Cosine & Warm-up w. 3ep \\
    EfficientNet-B0 & 50 & 128 & 1e-1$\to$1e-3  & SGD & 0.9 & 1e-4 & Cosine & Warm-up w. 3ep \\
    ViT-T & 50 & 512 & 1e-3$\to$1e-5  & Adamw & - & 5e-2 & Cosine & Warm-up w. 3ep \\
    Swin-T & 50 & 512 & 1e-3$\to$1e-5 & Adamw & - & 5e-2 & Cosine & Warm-up w. 3ep  \\
    \hdashline
    CIA-Net & 30 & 4 & 2e-5$\to$4e-6 & Adam & - & - & Step & decay 0.95 per ep\\ 
    \bottomrule
    \end{tabular}

\end{table}

\begin{table}[h]
%
\centering
\caption{Model complexity comparison in terms of number of trainable parameters and floating point operations per second (FLOPS).}\label{tab:network}
    \begin{tabular}{l|c|c} 
    \toprule
    Networks & Number of Parameters ($\times 10^6$) & FLOPS ($\times 10^8$)\\
    \hline 
    ResNet18 & 11.69 & 1.82\\
    ResNet50 & 25.56 & 4.11 \\
    MobileNetV3-Small & 2.54 & 59.66 \\
    EfficientNetB0 & 5.29 & 401.68 \\
    ViT-Tiny & 5.67 & 1.08 \\
    SwinTransformer-Tiny & 27.40 & 4.35 \\
    \hdashline
    CIA-Net & 15.67 & 331.27 \\
    \bottomrule
    \end{tabular}
\end{table}
%
%
\begin{table}[h]
\centering
\caption{Patch-level classification performance comparison on BACH
in terms of test top-1 accuracy and AUC. The Data, annotations, and description of BACH is freely available at \url{https://iciar2018-challenge.grand-challenge.org/}. For all the compared methods, we follow the same settings as Table 2 in the paper. We use ResNet-18 as the backbone. }\label{tab:cls_bach}
\resizebox{\linewidth}{!}{ 
    \begin{tabular}{l|ccccccc|c} 
    \toprule
    Metrics& Baseline & SA+LAB & SA+HED & SA+HSV & SN+LAB & SN+HED & SN+HSV & Ours\\
    \hline
    Accuracy & 72.68 & 74.18 & 74.22 & 73.43 & 74.98 & 62.50 & 74.01 & \textbf{79.11} \\
    AUC & 0.914 & 0.931 & 0.929 & 0.918 & 0.933 & 0.870 & 0.923 & \textbf{0.951} \\
    \bottomrule
    \end{tabular}
}
\end{table}

\begin{table}[h]
\caption{Classification comparison on BACH with MobileNetV3-Small. }\label{tab:cls_bach2}
\resizebox{\linewidth}{!}{ 
    \begin{tabular}{l|ccccccc|c} 
    \toprule
    Metrics& Baseline & SA+LAB & SA+HED & SA+HSV & SN+LAB & SN+HED & SN+HSV & Ours\\
    \hline
    Accuracy & 74.22 & 77.66 & 75.99 & 77.33 & 78.34 & 68.72 & 74.69 & \textbf{79.08} \\
    AUC & 0.925 & 0.937 & 0.937 & 0.938 & 0.944 & 0.884 & 0.924 & \textbf{0.949} \\
    \bottomrule
    \end{tabular}
}
\end{table}

\begin{figure*}[htbp]
\centering
\subfigure[]{
\begin{minipage}[t]{\tsnewidth\linewidth}
\centering
\includegraphics[width=1\linewidth]{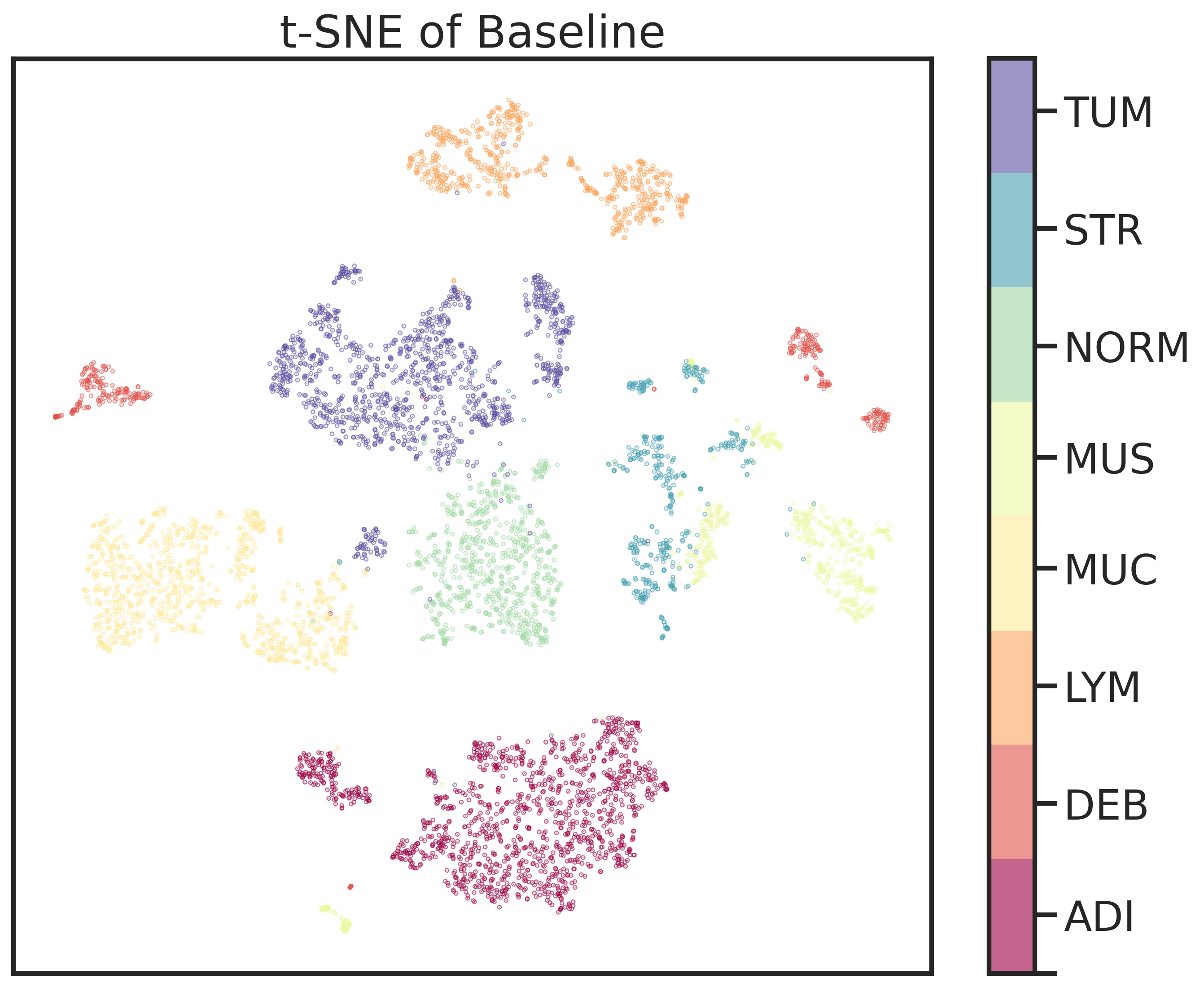} \label{fig:tsne-baseline}
\end{minipage}
}
%
\subfigure[]{
\begin{minipage}[t]{\tsnewidth\linewidth}
\centering
\includegraphics[width=1\linewidth]{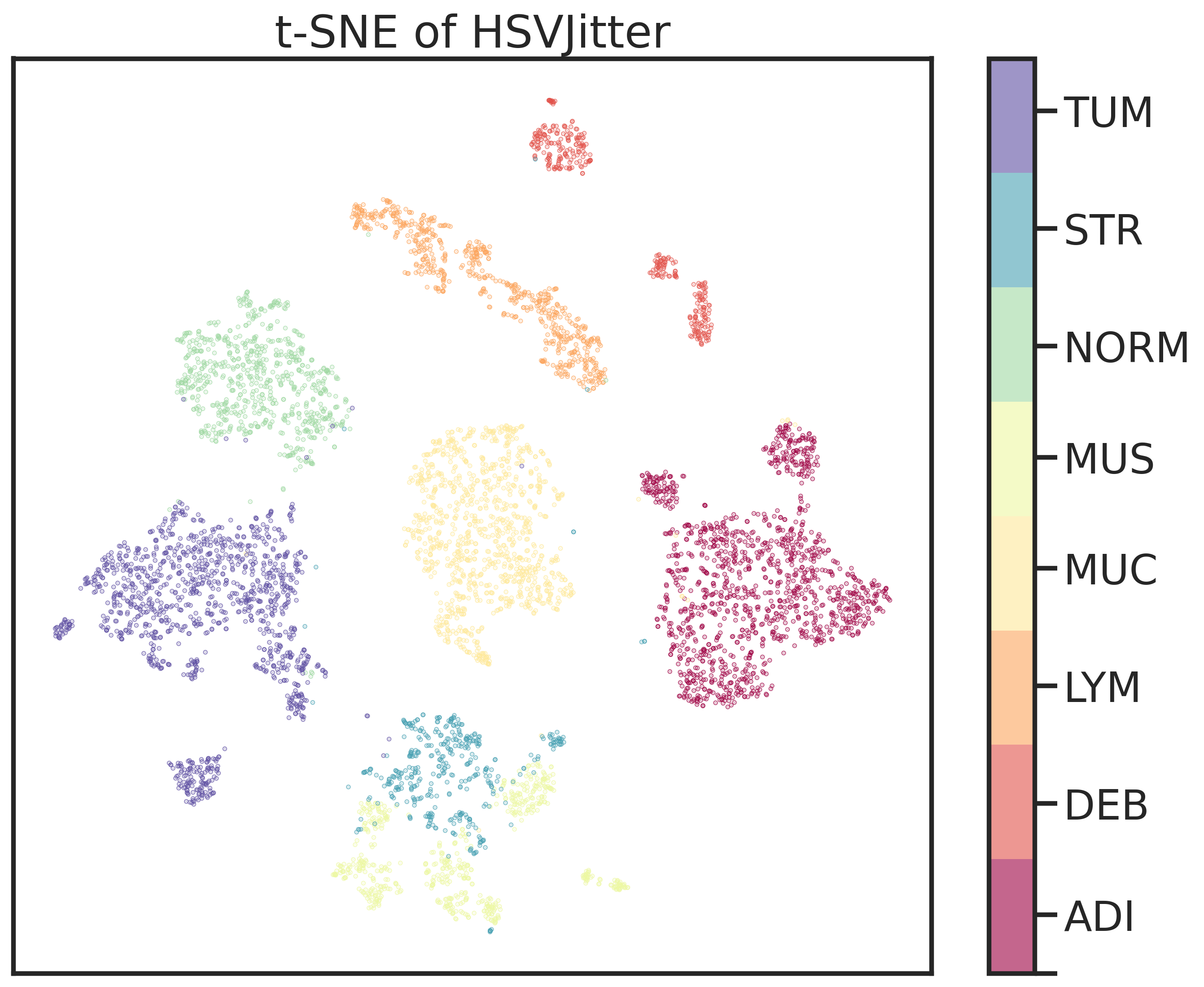} \label{fig:tsne-sn}
\end{minipage}
}
%
\subfigure[]{
\begin{minipage}[t]{\tsnewidth\linewidth}
\centering
\includegraphics[width=1\linewidth]{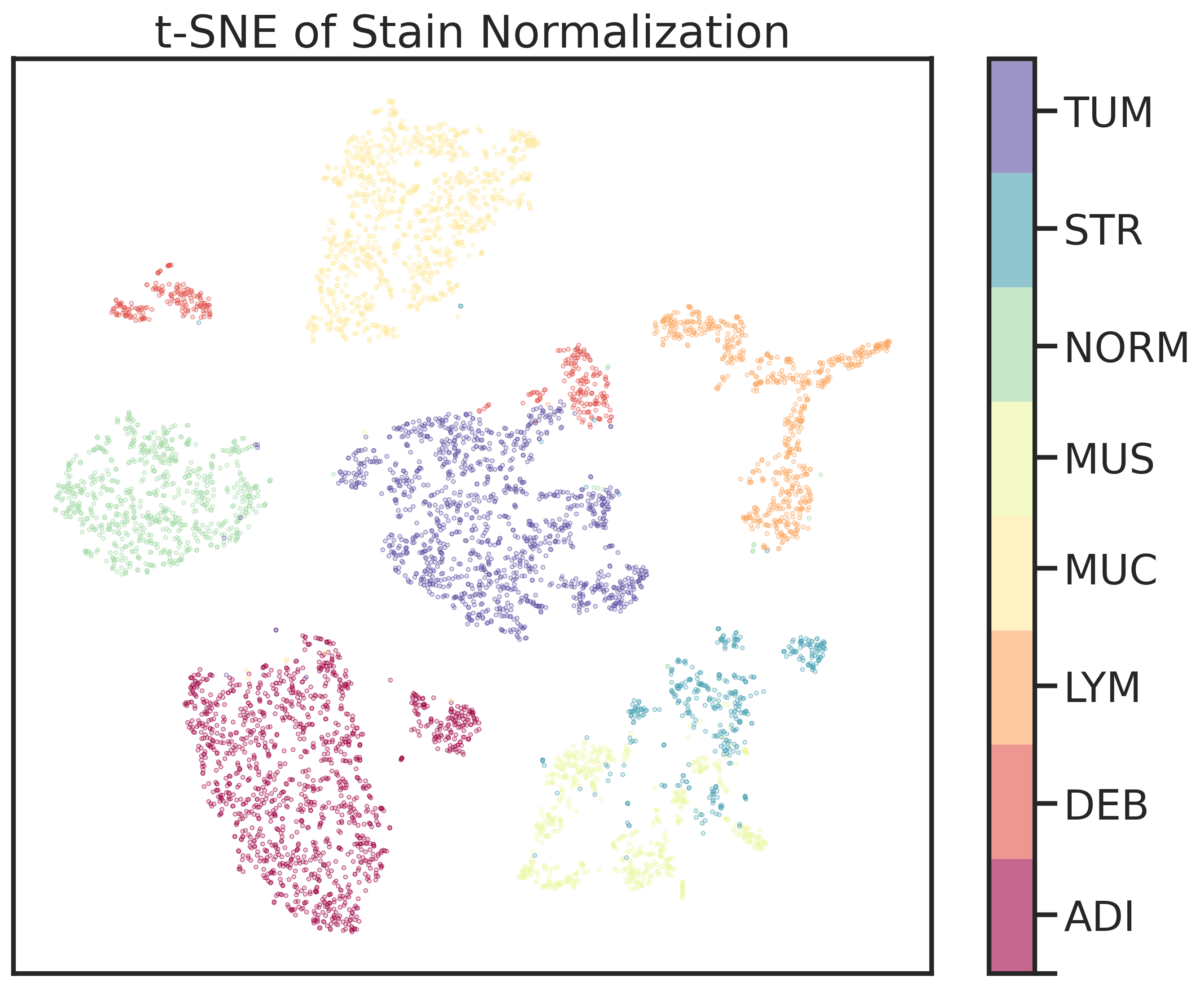} \label{fig:tsne-sa}
\end{minipage}
}
\subfigure[]{
\begin{minipage}[t]{\tsnewidth\linewidth}
\centering
\includegraphics[width=1\linewidth]{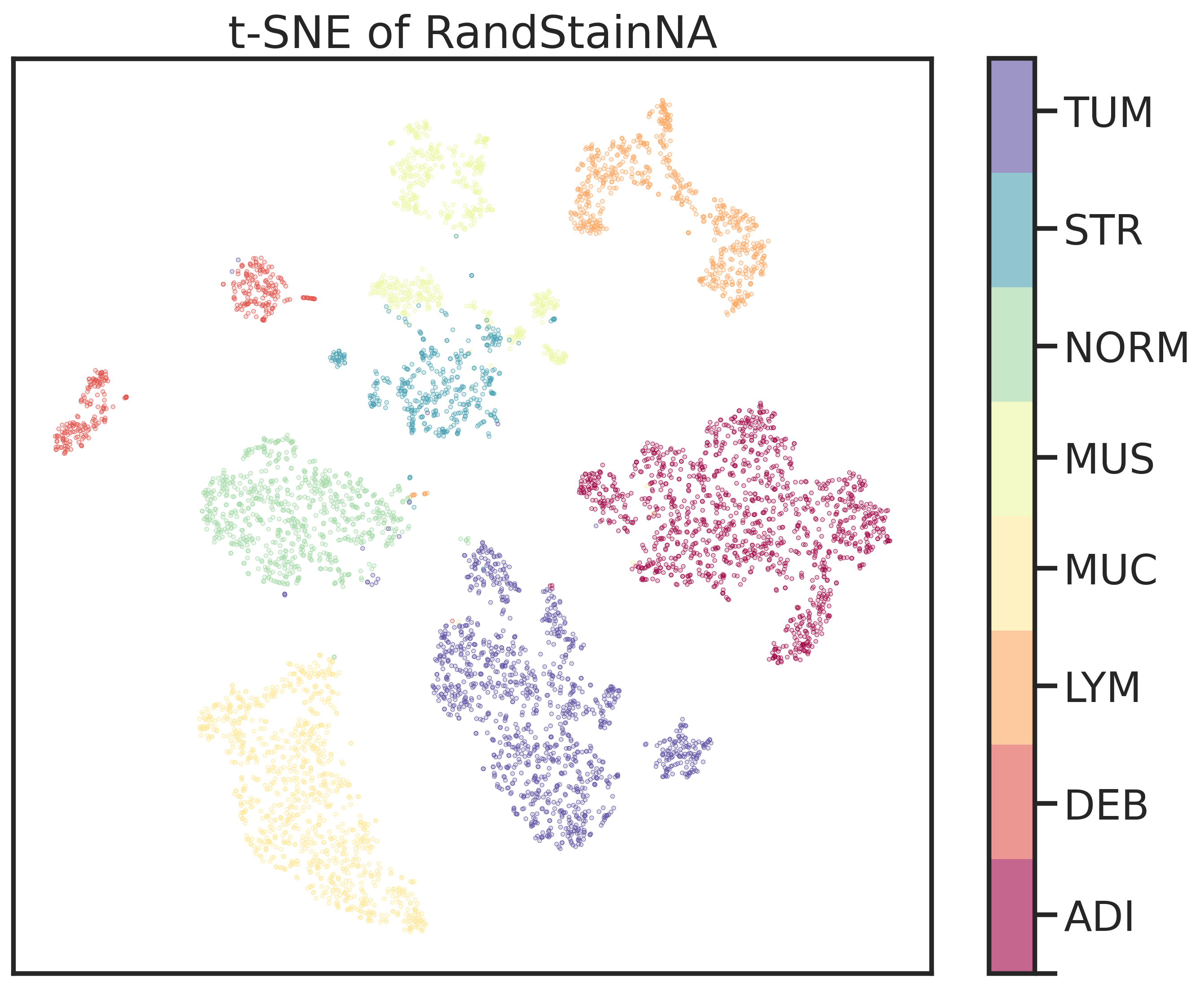} \label{fig:tsne-ours}
\end{minipage}
}
\caption{T-sne visualization of the penultimate layers trained with baseline (a), stain augmentation (b), stain normalization (c), our proposed RandStainNA (d).}
\label{fig:tsne}
\end{figure*}

\end{document}